\documentclass[twoside]{article}

\usepackage[accepted]{aistats2021}

\usepackage{microtype}
\usepackage{graphicx}

\usepackage{booktabs} %
\usepackage{xcolor}
 
\usepackage{mathabx}
\usepackage{changepage}
\usepackage{savesym}
\usepackage{caption}
\usepackage{subcaption}

\usepackage{amsmath} 
\usepackage{amsthm}
\usepackage{mathtools}
\usepackage{float}
\usepackage{bm}
\usepackage{dsfont}
\usepackage{graphicx,graphics}
\usepackage[inline]{enumitem}

\usepackage{amsthm}

\usepackage{multirow}

\DeclarePairedDelimiter\floor{\lfloor}{\rfloor}

\newtheorem{theorem}{Theorem}
\newtheorem{definition}{Definition}
\newtheorem{lemma}{Lemma}
\newtheorem{corollary}{Corollary}[theorem]

\begin{document}

\twocolumn[

\aistatstitle{Integer Programming-based Error-Correcting Output Code Design for Robust Classification}

\aistatsauthor{ Samarth Gupta \And Saurabh Amin}

\aistatsaddress{  samarthg@mit.edu \\ Massachusetts Institute of Technology  \And amins@mit.edu \\  Massachusetts Institute of Technology  }
]

\begin{abstract}
Error-Correcting Output Codes (ECOCs) offer a principled approach for combining simple binary classifiers into multiclass classifiers. In this paper, we investigate the problem of 
designing optimal ECOCs to achieve both nominal and adversarial accuracy using Support Vector Machines (SVMs) and binary deep learning models. In contrast to previous literature, we present an Integer Programming (IP) formulation to design minimal codebooks with
desirable error correcting properties. Our work leverages the advances in IP solvers to generate codebooks with optimality guarantees. To achieve tractability, we exploit the underlying graph-theoretic structure of the constraint set in our IP formulation. This enables us to use edge clique covers to substantially reduce the constraint set. Our codebooks achieve a high nominal accuracy relative to standard codebooks (e.g., one-vs-all, one-vs-one, and dense/sparse codes). We also estimate the adversarial accuracy of our ECOC-based classifiers in a white-box setting. Our IP-generated codebooks provide non-trivial robustness to adversarial perturbations even without any adversarial training.
\end{abstract}

\section{Introduction}
\vspace{-0.2cm}
Error Correcting Output Codes (ECOCs) offer an effective and flexible tool to combine individually trained binary classifiers for multiclass classification. Prior research \cite{ecoc,allwein} has shown that ECOCs can provide high multiclass classification accuracy using simple but powerful binary classifiers (e.g., Support Vector Machines and Adaboost). On the other hand, extensive body of work has emerged in recent years showing that, when large amount of training data is available, deep learning models \cite{DL_nature} outperform most multiclass classifiers. Still, further progress is needed for classification tasks when training data is limited or constrained, and model interpretability is preferred. In this paper, we consider the problem of ECOC-based multiclass classification, when individual binary classifiers are SVMs or deep learning models. We focus on the question of design of codebooks -- along with optimality guarantees.

Importantly, our approach to codebook design is distinct from the prior literature, which approaches the problem using a continuous relaxation of the inherently discrete optimization problem, and  solving the relaxed problem using nonlinear optimization tools~\cite{ crammer2002,sparse_ecoc,spectral_ecoc,ecf}. In principle, this approach can be scaled to a large number of classes, but it does not provide any optimality guarantees. Another approach in the literature \cite{ecoc} casts the design problem as a propositional satisfiability problem that can be solved for using off-the-shelf SAT solvers. However, only a feasible solution may be readily computable using this approach. In contrast, we formulate the optimal codebook design problem as a large-scale Integer Program (IP), and exploit the structure of the problem to obtain a compact formulation that can be solved with modern IP solvers. Our resulting codebook has optimality guarantees. This also enables a systematic comparison with respect to several well-known fixed-size codebooks. 

Our IP formulation is flexible in that it models various codebook (or coding matrix) generation criteria: (i) Sufficiently large Hamming distance between any pair of codewords ({\em row separation}); (ii) Uncorrelated columns ({\em column separation}); (iii) Relatively even distribution of data points across two classes ({\em balanced columns}); and (iv) Larger Hamming distance between pair of codewords whose corresponding classes are hard to separate from one another. These criteria are important not only for the nominal error correction performance, but also promote  adversarial robustness. However, this initial formulation can quickly become intractable for a classification problem of more than 10 classes. 

To address the abovementioned computational bottleneck, we exploit the inherent graph-theoretic feature of the constraints that pertain to selecting an appropriate subset of columns. In particular, we prove that the constraints modeling the pair of columns that do not satisfy column separation criterion can be replaced by a much smaller set formed by an {\em edge clique cover} of the underlying graph. This result allows us to reformulate our original problem into another IP with a substantially smaller set of constraints. %

A distinct advantage of our design approach is that it generates relatively small codebooks, which achieve a high nominal accuracy as well as robustness to adversarial perturbations \cite{intriguing,fgsm,one-pixel}. In particular, we demonstrate that our IP-generated codebooks outperform the well-known codebooks such as one-vs-all and one-vs-one, and other dense or sparse designs. To evaluate the robustness of optimal codebooks to adversarial perturbations, we conduct experiments based on white-box attacks \cite{MadryNet,tramer}. Importantly, our codebooks achieve non-trivial robustness even without any adversarial training of the individual binary classifiers. Thus, our results suggest a strong potential of ECOCs for training robust classifiers.

The paper is organized as follows: Sec.~\ref{sec:ecc} introduces the ECOC framework; Sec.~\ref{sec:criteria} presents the design criteria for codebooks; Sec. \ref{sec:ip_form} details our IP formulation; Sec. \ref{sec:exp} provides computational experiments on numerous datasets; Sec.~\ref{sec:conclude} outlines future work.
\section{ECOCs for Classification} \label{sec:ecc}
\begin{figure}[h]
  \begin{adjustwidth}{-0.1in}{-0.2in}  
    \begin{subfigure}{0.4\linewidth}
        \includegraphics[scale=0.3]{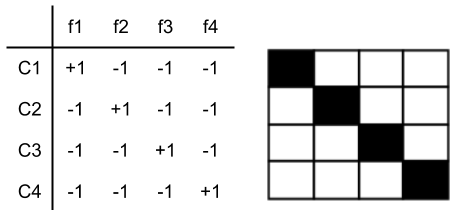}
        \caption{one-vs-all (binary)}\label{fig:ova}
    \end{subfigure}    
    \begin{subfigure}{0.45\linewidth} 
        \includegraphics[scale=0.3]{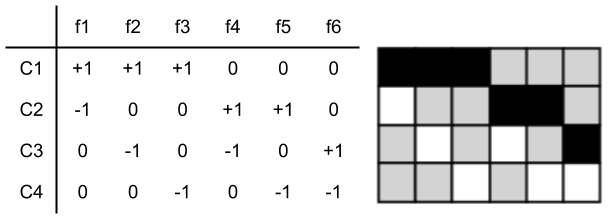}
        \caption{one-vs-one (ternary)}\label{fig:ovo}
    \end{subfigure}
    \end{adjustwidth}
    \caption{Examples of codebooks for a 4-class problem.} \label{fig:4-class}
\end{figure}

In the ECOC-based framework for $k$-class classification~\cite{ecoc}, each class is \emph{encoded} with a unique {codeword} of length $l$, resulting in a codebook (coding matrix) $ \mathcal{M}=(m_{ij})$ of size ${k\times l}$. For binary (resp. ternary) codes, the entries $m_{ij}$ of the coding matrix $\mathcal M$ belong to the set $\{+1,-1\}$ (resp. $\{+1,0, -1\}$). The rows (resp. columns) of $\mathcal{M}$ correspond to distinct classes (resp. binary classifiers or \emph{hypotheses}). Figure \ref{fig:4-class} shows examples of two standard codebooks. 

In the learning problem corresponding to every column in $\mathcal{M}$, the set of training examples belonging to different classes $C_1,\dots, C_k$ is partitioned into two groups: all examples from classes with entry $+1$ represent the positive class, and all examples from classes with entry $-1$ represent the other class. In the case of ternary codes, training examples with entry ${0}$ are not included in the training set and are considered irrelevant. 

Let  $f_1(\cdot),\dots, f_l(\cdot)$ represent the learned binary hypotheses for the corresponding columns of $\mathcal{M}$. 
For a learned hypothesis~$s \in \{1,\dots l\}$ and a test example~$x$, let $f_{s_{+1}}(x) $ (resp. $f_{s_{-1}}(x) $) denote the output/score of the class $+1$ (resp. class $-1$). Then,
\begin{equation*}
  f_s(x) :=
    \begin{cases}
      +1 & \text{if $f_{s_{+1}}(x)  > f_{s_{-1}}(x) $ }\\
      -1 & \text{otherwise}      
    \end{cases}    \quad   \forall s \in\{1,\dots,l\}.
\end{equation*}
After evaluating $x$ on all the $l$ hypotheses, we obtain an encoding $\vec{f}(x)= [ f_1(x),\dots, f_l(x)]$. To associate $\vec{f}(x)$ with a class (i.e., a row of coding matrix $\mathcal{M}$), we can use a decoding scheme based on a similarity measure such as Hamming distance. Particularly, one can compute the Hamming distance $d_H(\cdot,\cdot)$ between $\vec{f}(x)$ and each codeword $\mathcal{M}(r,\cdot)$ and select the class, denoted $\hat{y}$, that corresponds to the minimum distance:
\begin{align}
 &d_H(\mathcal{M}(r,\cdot), \vec{f}(x))  = \sum_{s=1}^{l} \bigg(\frac{1-\mathcal{M}(r,s) \times f_s(x)}{2} \bigg)  \label{eq:decode_hamming}\\
 &\hat{y} =\;  \underset{r}{\mathrm{argmin}} \;\; d_H(\mathcal{M}(r,\cdot), \vec{f}(x)). \label{eq:decode_class} 
\end{align}
\iffalse
It is important to note that the accuracy of multi-class classification scheme introduced above crucially depends on the error-correcting ability of the coding matrix $\mathcal{M}$. Thus, the coding matrix must be chosen carefully to ensure low test error. It is well-known that if every pair of distinct codewords (or rows) of the coding matrix $\mathcal{M}$ has a hamming distance of at-least $d$, then such a code can correct \textit{at-least $\floor*{\frac{d-1}{2}}$ errors}. Next we discuss our design criteria for generating a ``good'' codebook. 
\fi
\section{Codebook Generation Criteria}\label{sec:criteria}
The final prediction accuracy of ECOC scheme introduced in Sec.~\ref{sec:ecc}  crucially depends on the error correction ability of the coding matrix $\mathcal{M}$. To ensure low test error, the coding matrix must be chosen carefully. Below we introduce the key properties that serve as guidelines for our design of \emph{binary codes}.\footnote{Similar codebook design approach can be developed for ternary codes (not presented here due to space constraints).}

\textbf{Row Separation:}
It is well-known that more separation between pairs of codewords (i.e., rows in the coding matrix $\mathcal{M}$) improves the error correction capability. Particularly, if every pair of distinct codewords has a hamming distance of at-least $d$, then such a code can correct \textit{at-least $\floor*{\frac{d-1}{2}}$ errors}. Thus, we seek coding matrix with a high row separation between any pair of codewords.

\textbf{Column Separation:} 
Additionally, every pair of distinct columns in $\mathcal{M}$ should be uncorrelated.
The benefit of large column separation can be understood by drawing analogy with error correction in communication over a noisy channel. {Encoding} a signal and transmitting the codeword over a noisy channel is highly effective when the errors introduced during transmission are \textit{random}. By maintaining a sufficiently large encoding, one can recover the original signal at the receiving end with high accuracy. Analogously, in our setup, if any two columns (classifiers) make errors in their predictions on the same inputs (i.e., their outputs are highly correlated), then the effectiveness of encoding in correcting errors will be reduced. %

\vspace{0.2cm}

\textbf{Balanced Columns:}
On the other hand, to prevent over-fitting  of individual hypotheses, it is important to  prioritize selection of columns for which the $k$ class data points are evenly distributed across the two classes. This criterion is particularly relevant when the test examples are adversarially perturbed~\cite{trade-off}.

\vspace{0.2cm}

\textbf{Data Distribution:}
Finally, in multi-class problems, some class pairs are more difficult to separate than others. This makes the prediction of these classes more vulnerable to adversarial attacks. Therefore, it is desirable to have larger Hamming distances among pairs of codewords  corresponding to hard-to-separate class pairs. This hardness of separation can be estimated from the training data for different class pairs, either using the semantics of classes~\cite{sparse_ecoc}, or by calculating similarity measures between classes for small datasets ~\cite{spectral_ecoc,ecf,pujol1,hierarchial_koller,hierarchial_preona}.

\section{Integer Programming Formulation} \label{sec:ip_form}
In this section, we embed the abovementioned guidelines into a discrete optimization formulation for generating an optimal codebook. 

To begin with, note that for a $k-$class problem a coding matrix can have at most $({2^k} -  2)/2 = 2^{k-1} -1$ columns. However, such an exhaustive coding might be feasible only for a small $k$ (2 to 5). As $k$ increases, the number of binary classifiers that need to be trained for exhaustive coding increase exponentially. Practically, it is desirable to select a small subset (say of $L$ columns) from $2^{(k-1)}-1$ possible columns. This subset should be selected in accordance with the codebook generation criteria described in Sec.~\ref{sec:criteria}.

One way to formulate the column subset selection problem is to cast it as a propositional satisfiability problem, and solve it using an off-the-shelf SAT solver. For example, the authors in~\cite{ecoc} considered the following problem for $8\le k \le 11$: For a predefined number of columns $L$ and some value $\rho$, is there a solution such that the Hamming distance between any two columns is between $\rho$ and $L-\rho$? However, this approach only leads to a feasible (not necessarily optimal) solution. In contrast, we present an Integer Programming (IP) problem that captures the design criteria in a more flexible manner and can be used to find an {\em optimal} codebook. 

For sake of simplicity, we first consider the row and column separation criteria; the remaining criteria on balanced columns and data distribution can be addressed in our IP formulation, as discussed subsequently at end of this section. In its basic form our problem is the following: We want to find a solution which \emph{maximizes the minimum Hamming distance between any two rows (or the error-correcting property).} 

Let $x_i$ denote the binary variable associated with each column $i$ of the exhaustive code for $i \in \{1,\dots 2^{k-1} -1\}$, i.e. the decision variable whether or not column $i$ is selected in the final solution. Also, let $x_{ij}$ be the binary variable which represents the outcome of AND operation between variables $x_i$ and $x_j$ for all distinct $i,j$ pairs, i.e.  $(i,j) \in \{1,\dots, 2^{k-1} -1\}^2| i < j$. Essentially, when $x_{ij}=1$ means that columns $i$ and $j$ satisfy the column separation criterion. We can now write the IP formulation to generate an optimal codebook as follows:
 \begin{adjustwidth}{-0.08in}{-0.2in}
\begin{alignat}{10}
 \bm{\mathcal{IP}1:}&  \max_{x_i,x_{ij}} \; \min\ \;  \{&&  d_H^{1,2}(x_i), \dots ,\; d_H^{k-1, k}(x_i) \;\}\\
	    & \text{s.t.}  \notag \\
	    & \sum_{i=1}^{2^{k-1} -1 }x_i \leq && L \\
 	    &  \rho\; x_{ij} \leq d_H \big( &&\mathcal{M}  (\cdot,i) ,\mathcal{M}(\cdot,j)\big)\; x_{ij} \leq (L-\rho)\;x_{ij}  \notag \\
 	    &          && \forall\; (i,j) \in \{1,\dots, 2^{k-1} -1\}^2 |\; i < j  \label{ip:column} \\
	    &  x_{ij} \leq x_i      \label{ip:and1}\\
	    &  x_{ij} \leq x_j    \label{ip:and2}\\
	    &  x_i + x_j -1 && \leq  x_{ij}  \label{ip:and3}\\
	    &  d_H^{s,t}(x_i)  = &&\sum_{i=1}^{2^{k-1} -1} \Big(\frac{1- \mathcal{M}(s,\cdot)\times\mathcal{M}(t,\cdot)  }{2}\Big) x_i  \notag \\
	    &  \hphantom{}   && \hphantom{} \forall\; (s,t) \in \{1, \dots, k \}^2 |\; s < t \\
	    &  x_i     \in \{0,  1 \} 				&& \forall\; i \in \{1, \dots, 2^{k-1} -1 \}\\
	    & x_{ij}  \in \{0,  1\}	&& \forall\; (i,j) \in \{1,\dots, 2^{k-1} -1\}^2 |\; i < j    
\end{alignat}
\end{adjustwidth}
\clearpage
In $\mathcal{IP}1$, max-min objective can be simplified by introducing an auxiliary variable $t$, where $ t = \min\ \;  \{d_H^{1,2}(x_i),\; d_H^{1,3}(x_i),\; \dots ,\; d_H^{k-1, k}(x_i) \;\}$, and adding the corresponding constraints $ t \leq d_H^{1,2}(x_i) \;,\;t\leq d_H^{1,3}(x_i) \;,\; \dots \;, t\leq d_H^{k-1,k}(x_i)$. Eq. \eqref{ip:column} ensures large column separation for $x_{ij}=1$. Constraints \eqref{ip:and1} and \eqref{ip:and2} ensure that if $x_{ij}=1$ then both columns $i$ and $j$ are included in the solution, i.e.  $x_i=1$ and $x_j=1$. Conversely, Equation \eqref{ip:and3} ensures that if columns $i$ and $j$ are selected then $x_{ij}=1$.

\vspace{0.2cm}

We note that in $\mathcal{IP}1$ there are $2^{k-1} -1 \approx \mathcal{O}(2^{k-1})$ binary variables for each column, and for each pair of columns there are $2^{k-1} -1 \choose 2$ $ \approx \mathcal{O}(2^{2k-3})$ binary variables. Thus, the total number of binary variables are $\mathcal{O}(2^{2k-3})$. Similarly, the total number of constraints are $\mathcal{O}(2^{2k-1})$. For $k=10$, this would entail solving an IP of approximately $130,000$ variables and $650,000$ constraints. Modern IP solvers like Gurobi and CPLEX can handle such problem instances. 

\vspace{0.2cm}

However, for $k>10$, the above optimization problem quickly becomes intractable. The main reason is that we have a binary variable~$x_{ij}$ for each pair of columns to capture the large column separation criterion; see~\eqref{ip:column}. We now propose a second formulation which does not involve a new variable for every pair of columns. 

\vspace{0.2cm}

Let $\mathcal{S}_p$ denote the set of all distinct pairs of columns in the exhaustive code $\mathcal{M}$, i.e. $ \mathcal{S}_p = \{(i,j) \in \{1,\dots, 2^{k-1} -1\}^2 |\; i < j\}$ and $ |\mathcal{S}_p| = \binom{2^{k-1} -1}{ 2}$. We now consider two mutually disjoint subsets $\mathcal{S}_p^{feas}$ and $\mathcal{S}_p^{inf}$, such that  $\mathcal{S}_p = \mathcal{S}_p^{feas} \cup \mathcal{S}_p^{inf}$: the set $\mathcal{S}_p^{feas}$ (resp. $\mathcal{S}_p^{inf}$) contains only those $i,j$ pairs that \textit{satisfy} (resp. \textit{do not} satisfy) the column separation criterion~\eqref{ip:column}. Mathematically, we can write:
\begin{alignat}{10}
& \mathcal{S}_p  && = \Big\{(i,j) \in \{1,\dots, 2^{k-1} -1\}^2 |\; i < j   \Big \},\\
& \mathcal{S}_p^{feas}  && = \Big\{(i,j) \in \{1,\dots, 2^{k-1} -1\}^2 |\; i < j \text{ and} \notag\\ 
&			 && \quad \quad  \rho \leq d_H\big(\mathcal{M}(\cdot,i),\mathcal{M}(\cdot,j)\big)\leq (L-\rho)    \Big \}, \\
& \mathcal{S}_p^{inf}  && = \mathcal{S}_p \setminus \mathcal{S}_p^{feas}.
\end{alignat}
In this new representation, the constraint \eqref{ip:column} is captured by the construction of $\mathcal{S}_p^{feas}$, which eliminates the need of variables $x_{ij}$ for column pairs. Similarly, we no longer need the constraints~\eqref{ip:and1}, \eqref{ip:and2} and \eqref{ip:and3}. Now, for any $(i,j)$ pair of columns in the set $\mathcal{S}_p^{inf}$, \textit{at-most} one of the columns can be included in the final solution. This can be achieved by setting $x_{ij} =0$ in~\eqref{ip:and3}. Equivalently, for every $(i,j)$ pair in $\mathcal{S}_p^{inf}$, it is sufficient to impose the constraint $x_i + x_j -1 \leq 0$.

We can now write $\mathcal{IP}1$ as the following equivalent form:
\begin{alignat}{10}
	 \bm{ \mathcal{IP}2:}& \max_{x_i} \; \min\ \; \{ && d_H^{1,2}(x_i),\dots , d_H^{k-1, k}(x_i) \;\} \notag\\
	 \text{s.t.}   & \sum_{i=1}^{2^{k-1} -1 }x_i \leq && L \notag \\
	    &  x_i + x_j  \leq 1 \hphantom{\text{}} && \hphantom{MMM}\forall\;  (i,j) \in \mathcal{S}_{p}^{inf}      \label{ip2:inf}\\
 	    &  d_H^{s,t}(x_i)  = &&\sum_{i=1}^{2^{k-1} -1}\Big( \frac{1- \mathcal{M}(s,\cdot) \times \mathcal{M}(t,\cdot)}{2} \Big)x_i \notag \\
 	    &  \hphantom{} && \hphantom{MMM} \forall\; (s,t) \in \{1, \dots, k \}^2 |\; s < t \notag\\
	    &  x_i  \in \{0,  1 \} && \hphantom{MMM} \forall\; i \in \{1, \dots, 2^{k-1} -1  \} \notag
\end{alignat}

Since $\mathcal{IP}2$ does not contain any $x_{ij}$ variables, this formulation has significantly less number of variables and constraints in comparison to~($\mathcal{IP}1$). The computational complexity of $\mathcal{IP}2$ is mainly governed by the size of~$\mathcal{S}_{p}^{inf}$, which determines the number of constraints in~\eqref{ip2:inf}. However, even in this new representation, the size of the set $\mathcal{S}_{p}^{inf}$ becomes prohibitively large as $k$ increases. Table \ref{tab:constr}, column 4 shows how quickly $|\mathcal{S}_p^{inf}|$ increases with $k$ for an appropriately chosen $\rho$.

\begin{figure}[h!]
  \centering
 \includegraphics[scale = 0.45]{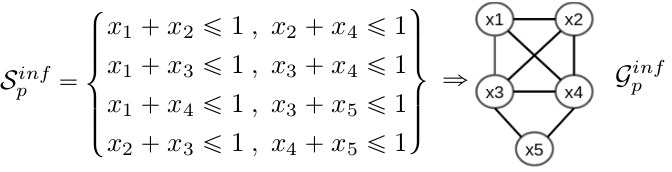}
 \caption{An example of $\mathcal{S}_p^{inf}$ and corresponding $\mathcal{G}_p^{inf}$.}\label{fig:setToGraph}
\end{figure}

Fortunately, the constraints~\eqref{ip2:inf} for the set $\mathcal{S}_p^{inf}$ can be represented on a graph $\mathcal{G}_p^{inf}$, in which each node corresponds to a column $x_i $ and each constraint $x_i + x_j \le 1$ corresponds to an edge between the node $i$ and node $j$; see Figure \ref{fig:setToGraph} for an illustration. 

In fact, the above graphical interpretation leads to a reduction in the number of constraints involving $(i,j)$ column pairs in the set $\mathcal{S}_p^{inf}$. Before presenting this result, we recall that a {\em clique} is a set of nodes in a graph such that there is an edge between any two distinct nodes of this set. Proofs of all results are provided in the supplementary material.

\begin{theorem}\label{th:clique}
The feasible space enclosed by the constraints constituting the edges of any clique $\mathcal{C} $ in $\mathcal{G}_p^{inf}$ is same as that enclosed by the single constraint:
\begin{equation}
  \centering
 \displaystyle \sum_{i \in \mathcal{C}} x_i \le 1. \label{eq:single_const}
\end{equation}
\end{theorem}

From theorem~\ref{th:clique}, we obtain that for a clique of size $n$, $n(n-1)/2$ constraints of form $x_i + x_j \leq 1 $  between all $(i,j)$ node pairs in the clique 
can be substituted with a single constraint \eqref{eq:single_const}. This constraint captures the requirement that out of all the columns in $\mathcal{M}$ forming a clique, at most one can be present in a feasible solution. Before introducing our next result, we recall the following useful definition.

\begin{definition}[Edge Clique Cover \cite{ecc,JensCovering, KouCovering}]
An edge clique cover for a graph $\mathcal{G}$, denoted as $\mathcal{ECC}(\mathcal{G})$,  is a set of cliques $\mathcal{ECC}(\mathcal{G}) =  \{ \mathcal{C}_1 , \mathcal{C}_2 ,\dots, \mathcal{C}_k \}$ such that:
\begin{enumerate}
 \item No clique $\mathcal{C}_i$ is contained in another clique $\mathcal{C}_j$, i.e $\mathcal{C}_i \nsubseteq \mathcal{C}_j$ for all $i \ne j$, and
 \item Every edge in the graph $\mathcal{G}$ is included in at-least one clique.  
\end{enumerate}
\end{definition}

\begin{corollary} \label{th:edge_cover}
The feasible space enclosed by the constraint set $\mathcal{S}_p^{inf}$ (or its graphical equivalent $\mathcal{G}_p^{inf}$)  in $\mathcal{IP}2$ is same as that enclosed by a much smaller constraint set formed by $\mathcal{ECC}(\mathcal{G}_p^{inf})$.
\end{corollary}
\iffalse
\emph{Proof:} Refer to Supplementary section.
\emph{Proof:} Since the graph $\mathcal{G}_p^{inf}$ does not contain any isolated nodes and loops, therefore we can write:
 $\bigcup_{i=1}^k \mathcal{C}_i = \mathcal{G}_p^{inf} $.
 Also, $\mathcal{G}_p^{inf} \equiv   \mathcal{S}_p^{inf}$, therefore $\{ \mathcal{C}_1, \dots ,\mathcal{C}_k\} \equiv \mathcal{S}_p^{inf}$. %
\fi
\begin{figure}[h!]
    \centering
\includegraphics[scale = 0.45]{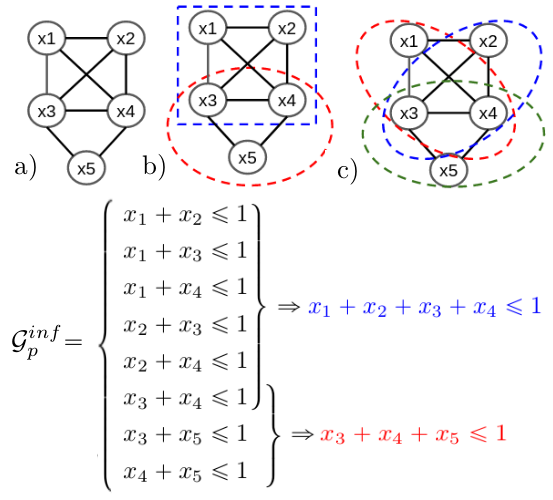}%
\caption{Graphical depiction of an example $\mathcal{S}_p^{inf}$ in (a), with two feasible edge clique covers ((b) and (c)). For edge-cover in (b), we show the reduced set of constraints corresponding to its cliques in blue and red. }\label{fig:clq}
\end{figure}
 
A given graph can have many possible edge clique covers; see for example  Fig.~\ref{fig:clq}. To reduce the size of the constraint set $\mathcal{S}_p^{inf}$ as much as possible, we would need to find an edge clique cover of the smallest size. However, the minimum edge cover problem is known to be NP-hard \cite{ecc_nphard}. Fortunately, several heuristics have been proposed to \cite{Kellerman,JensCovering, KouCovering, ecc} find edge clique cover of a graph, and they have been very effective in many practical applications. The heuristic \cite{ecc} is particularly well-suited for large graphs -- in practice, it shows a linear runtime in the number of edges. We therefore use this heuristic for our analysis.

We can further extend Collorary \ref{th:edge_cover} to generate edge-clique-covers of very large graphs in a distributed manner using the following result:
\begin{lemma} \label{lemma:1}
Suppose $ \mathcal{G}_1, \dots \mathcal{G}_m$ are edge-disjoint subgraphs of  $\mathcal{G}_p^{inf}$, such that:
\begin{enumerate}
 \item $ \mathcal{G}_i \cap  \mathcal{G}_j = \phi \;\; \forall \;  i,j \in \{1, \dots, m\}^2| i< j $
 \item $ \bigcup_{i=1}^m \mathcal{G}_i = \mathcal{G}_p^{inf}$
\end{enumerate}
The union of the edge clique covers of individual subgraphs $ \mathcal{G}_1, \dots \mathcal{G}_m$ is a valid edge clique cover of $\mathcal{G}_p^{inf}:$ $\;\; \bigcup_{i=1}^m \mathcal{ECC}(\mathcal{G}_i) = \mathcal{ECC}(\mathcal{G}_p^{inf})$. 
\end{lemma}
\begin{figure}[]
   \centering
\includegraphics[ scale = 0.44]{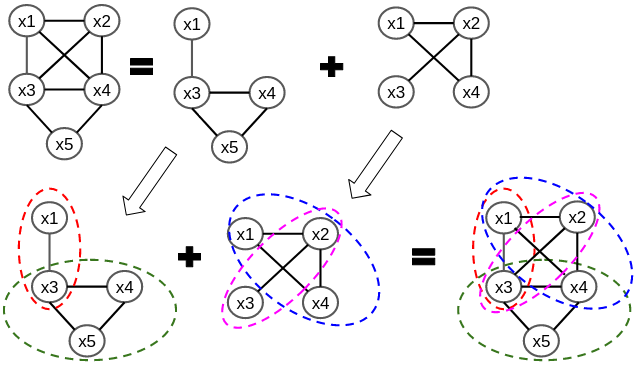}%
\caption{  Edge Clique Cover  generated by combining the edge clique covers of the individual subgraphs (Lemma \ref{lemma:1}). }\label{fig:clq_dist}
\end{figure}
Finally, using corollary \ref{th:edge_cover} (or its extension lemma \ref{lemma:1}) we can reduce $\mathcal{IP}2$ to the following integer program:
\begin{adjustwidth}{-0.1in}{-0.1in}
\begin{alignat}{10}
	\bm{ \mathcal{IP}3:}    & \max_{x_i} \; \min\  \{ &&d_H^{1,2}(x_i), \dots , d_H^{k-1, k}(x_i) \} \label{eq:ip3_obj}\\
	    & \text{s.t.} \notag \\
	    & \sum_{i=1}^{2^{k-1} -1 }x_i \leq && L \notag \\
	    &  \sum_{i: \forall i \in \mathcal{C}_{t}}x_i  \leq && 1 \hphantom{\text{}} \hphantom{MMM}\forall\;  \mathcal{C}_t \in \mathcal{ECC}(\mathcal{G}_{p}^{inf})      \label{ip1:and3}\\
 	    &  d_H^{s,t}(x_i) \;\; = && \sum_{i=1}^{2^{k-1} -1} \Big( \frac{1- \mathcal{M}(s,\cdot) \times \mathcal{M}(t,\cdot)}{2} \Big)x_i \notag \\
 	    &  \hphantom{} && \hphantom{MMM} \forall\; (s,t) \in \{1, \dots, k \}^2 |\; s < t \notag\\
	    &  x_i  \in \{0,  1 \} && \hphantom{MMM} \forall\; i \in \{1, \dots, 2^{k-1} -1  \} \notag
\end{alignat}
\end{adjustwidth}
\begin{table*}[h]
\centering
\caption{Reducing the size of the constraint set $|\mathcal{S}_p^{inf}|$ in $\mathcal{IP}2$ by finding the Edge Clique Cover of $\mathcal{G}_p^{inf}$.}
\resizebox{0.9\linewidth}{!}{%
\begin{tabular}{|c|c|c|c|c|c|c|c|}
\hline
\begin{tabular}[c]{@{}c@{}}No. of classes \\ $\bm{k}$ \end{tabular} & \begin{tabular}[c]{@{}c@{}}No. of Columns \\ $\bm{2^{k-1}-1}$\end{tabular}& $\rho$ &  \begin{tabular}[c]{@{}c@{}} No. of constraints\\ $\bm{|\mathcal{S}_p^{inf}|}$  \end{tabular} & \begin{tabular}[c]{@{}c@{}} No. of constraints \\ \textbf{(Reduced)} \end{tabular} & \begin{tabular}[c]{@{}c@{}} Reduction \\ Factor \end{tabular} & \begin{tabular}[c]{@{}c@{}}Time Taken\\ (in sec.)\end{tabular} \\ \hline
10 & 511      & 3 & 11,475        & 695        & \textbf{16}  & 0.146 \\ \hline 
11 & 1,023    & 3 & 28,105        & 1,404      & \textbf{20}  & 0.208 \\ \hline 
12 & 2,047    & 4 & 236,313       & 8,165      & \textbf{28}  & 0.991 \\ \hline 
13 & 4,095    & 4 & 610,006       & 18,472     & \textbf{33}  & 2.573 \\ \hline 
14 & 8,191    & 4 & 1,543,815     & 41,088     & \textbf{37}  & 7.390 \\ \hline 
15 & 16,383   & 5 & 12,040,770    & 44,916     & \textbf{268} & 58.957 \\ \hline 
16 & 32,767   & 5 & 31,783,020    & 91,304     & \textbf{348} & 249.53 \\ \hline 
17 & 65,535   & 5 & 82,441,772    & 185,661    & \textbf{444} & 935.76 \\ \hline
18 & 131,071  & 6 & 616,094,535   &  1,073,248 & \textbf{574} & 10075.8 \\ \hline
18 & 131,071  & 6 & 616,094,535   & \begin{tabular}[c]{@{}c@{}}5,952,906 $+$ 622,604  \\ $=$ 6,575,510 \end{tabular} & 93 & \begin{tabular}[c]{@{}c@{}}16251.667 $+$ 4977.376 \\ $=$ 21229.04 \end{tabular} \\ \hline
\end{tabular}
}\label{tab:constr}
\end{table*} 

Finally, the last two criteria mentioned in Sec.~\ref{sec:criteria} can be easily incorporated in $\mathcal{IP}3$. Specifically, the requirement for balanced columns can be incorporated by setting the $x_i$'s violating this criterion to $0$ in $\mathcal{IP}3$. Equivalently, since each $x_i \in \{0,1\}$ corresponds to whether a column is selected from the exhaustive code $\mathcal{M}$, we can simply reduce $\mathcal{M}$ by removing the unbalanced columns and then form $\mathcal{IP}3$. In contrast to \cite{spectral_ecoc}, in our formulation, the requirement for balanced columns further reduces the final problem size and complexity.

The remaining criterion of data distribution can be also incorporated by modifying the objective function. Previous works such as \cite{ecf, sparse_ecoc, spectral_ecoc}, pre-compute a similarity measure between every pair of classes (from training data) and use this computation to estimate the desirable class-pairwise hamming distances  $\hat{d}_{p,q}$. Finally, they optimize to obtain codebooks which attain these distance values. This can be easily incorporated in our formulation by changing the objective function \eqref{eq:ip3_obj} in $\mathcal{IP}3$ to the following: 
\begin{equation}
 \centering
\displaystyle \min_{x_i} \; \sum_{ (p,q) \in \{1, \dots, k\}^2| p < q} | d_H^{p, q}(x_i)- \hat{d}_{p,q}|. \label{ip3:new_obj}
\end{equation}
\iffalse
We would like to point out that in case if the IP solver does not terminate for larger $k$, i.e. $k>11$, it will still provide us with a feasible solution and moreover it will also provide us with an upper-bound on our objective function value. This will provide us with some idea on how far we are from an optimal solution if a feasible solution with the same value as that of upper-bound exists. This is another major benefit of our IP based formulation. %
\fi
\section{Experiments}\label{sec:exp}
We run all our experiments on a system with a single 1080Ti Nvidia GPU, Intel Core i7-6800K CPU and 128 GB RAM. We use Gurobi as our IP solver.
\begin{table}[ht]
\centering
\caption{ $\mathcal{IP}3$: Optimality Gap (max. time 2000s).}
\resizebox{0.8\linewidth}{!}{%
\begin{tabular}{|c|c|c|c|c|c|}
\hline
$k$  & $L$  & $f_{\text{best}}$ & \begin{tabular}[c]{@{}c@{}}Best\\ Bound\end{tabular} & Gap & \begin{tabular}[c]{@{}c@{}}Optimality Gap\\ $|f_{\text{best}} - f^*|$\end{tabular}  \\ \hline
10 & 20 & 10                                                       & 10            & 0\% & $\bm{0\%}$                                                                                  \\ \hline
11 & 22 & 12                                                       & 12            & 0\% & $\bm{0\%}$                                                                                  \\ \hline
12 & 24 & 12                                                       & 12            & 0\% & $\bm{0\%}$                                                                                  \\ \hline
13 & 26 &  13                                                      & 14            &7.69\%     &   $\bm{7.7\%}$                                                                             \\ \hline
14 & 28 &  14                                                      & 15            & 7.14\%    &   $\bm{0\%}$                                                                             \\ \hline
15 & 30 &  15                                                      & 16            & 6.67\%    &   $\bm{0\%}$                                                                             \\ \hline
16 & 32 &  16                                                      & 17            & 6.25\%    &   $\bm{0\%}$                                                                             \\ \hline
17 & 34 &  16                                                      & 18            & 12.2\%     &  $\bm{6.25\%}$                                                                            \\ \hline
18 & 36 &  17                                                      & 19            &  11.8\%   &   $\bm{5.5\%}$                                                                             \\ \hline
\end{tabular}
}\label{tab:gap}
\end{table}

Our computational experiments focus on solving $\mathcal{IP}3$ which uses the edge-clique-cover approach to reduce the constraint set $\mathcal{S}_p^{inf}$. Table \ref{tab:constr} shows the reduction in size of set $\mathcal{S}_p^{inf}$ as the number of classes~$k$ increases. Notably, for $k \ge 15$ we achieve a reduction of more than \emph{two orders of magnitude}, which demonstrates the advantage of using our approach. The last row shows the performance of generating the edge-cover on two different subgraphs obtained after partitioning the original graph, thus validating the lemma~\ref{lemma:1}.

Thanks to the reduced constraint set, we can solve $\mathcal{IP}3$ and obtain the optimality gap for different instances as shown in Table \ref{tab:gap}. $f_{best}$ denotes the objective function value of the best solution and ``Best Bound'' denotes the best upper bound found by Gurobi. We obtain an optimal solution or a relatively small optimality gap.  Thus, our formulation is tight and enables Gurobi to terminate quickly without exploring a large branch-and-bound tree. These results demonstrate our approach to codebook design indeed provides \emph{low optimality gaps.} 

We now evaluate the classification performance of our IP-generated codebooks in both natural and adversarial settings. We compare performance against various standard codebooks: 1-vs-all \cite{ova} and 1-vs-1 as well as Sparse and Dense codes generated using the procedure outlined in \cite{allwein}.\footnote{Please see the supplementary section for more details.}%

\begin{figure*}[]
 \centering
  \begin{adjustwidth}{-0in}{-0in}  
  \centering
    \begin{subfigure}{0.32\linewidth}
     \includegraphics[scale = 0.4]{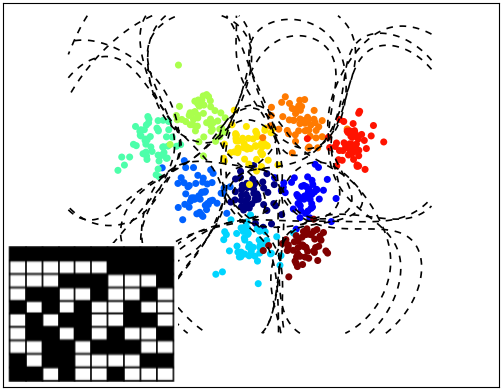}
     \caption{$\mathcal{IP}3$ generated ($89.8\% $)}
    \end{subfigure}
   \begin{subfigure}{0.32\linewidth}
     \includegraphics[scale = 0.4]{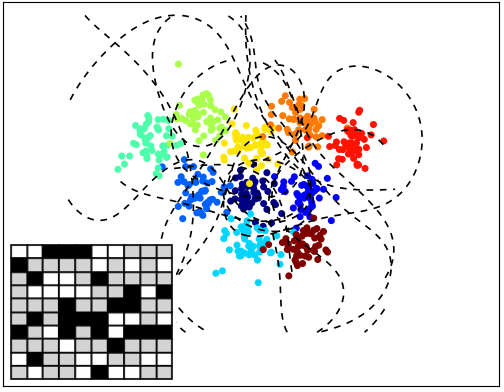}
     \caption{Sparse ( $66.8\% $)}
    \end{subfigure} 
     \begin{subfigure}{0.32\linewidth}
      \includegraphics[scale = 0.4]{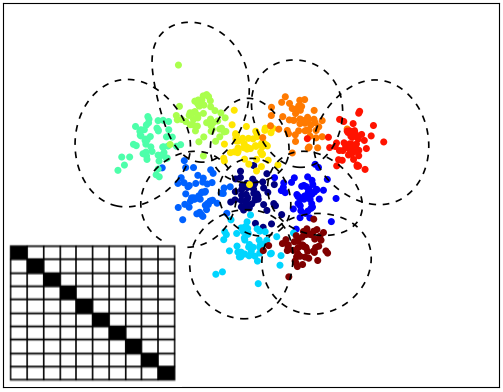}
      \caption{1-vs-All ( $80.6\% $)}
     \end{subfigure}
    \caption{Decision boundaries of different hypotheses in three different codebooks on 2d dataset.} \label{fig:2d}
 \end{adjustwidth}
\end{figure*}

\subsection{Natural Classification Performance}
\textbf{Toy Dataset (2d):} We generate a synthetic dataset of $10$ classes where points in each class are sampled from a 2d Gaussian distribution. Here we use SVMs with Rbf kernels as our binary classifier for individual hypotheses in all our codebooks. Figure~ \ref{fig:2d} shows the decision boundaries of all hypotheses for three codebooks along with the training set. The prediction accuracy on the test set is reported in Table \ref{tab:2d}. Our $\mathcal{IP}3$ generated codebook easily outperforms other codebooks, and almost matches the accuracy of  1-vs-1. Note that this codebook only used $L=20$ columns while 1-vs-1 used $L=45$ columns. This highlights the benefit of ECOC theory: high accuracy can be achieved with a carefully chosen \emph{compact} code-book.
\begin{table}[h]
\centering
\caption{Performance  on  2d Toy dataset ($k=10$).}
 \resizebox{\linewidth}{!}{%
\begin{tabular}{|c|c|c|c|c|c|}
\hline
\multicolumn{2}{|c|}{$\mathcal{IP}3$} & \multirow{2}{*}{\begin{tabular}[c]{@{}c@{}}Dense\\ $L=10$\end{tabular}} & \multirow{2}{*}{\begin{tabular}[c]{@{}c@{}}Sparse\\ $L=10$\end{tabular}} & \multirow{2}{*}{\begin{tabular}[c]{@{}c@{}}1-vs-All\\ $L=10$\end{tabular}} & \multirow{2}{*}{\begin{tabular}[c]{@{}c@{}}1-vs-1\\ $L=45$\end{tabular}} \\ \cline{1-2}
$L=10$             & $L=20$            &                                                                         &                                                                          &                                                                            &                                                                            \\ \hline
$89.8\%$             & $90.8\%$            & $88.1\%$                                                                    & $66.8\%$                                                                     & $80.6\%$                                                                       & $91.2\%$                                                                       \\ \hline
\end{tabular}
}\label{tab:2d}
\end{table}

\textbf{Real-world Datasets (Small/Medium): }
We evaluate the performance of different codebooks on small to medium sized, real-world datasets. We consider {\em Glass}, {\em Ecoli} and {\em Yeast} datasets taken from UCI repository \cite{uci}.
Details such as the number of samples, features and classes for each dataset are provided in the supplementary material.
For Dense, Sparse, and IP generated codebook we set $L = 2k$.
We again use SVMs with Rbf kernel as the binary classifier for training different hypotheses in our IP-generated and other codebooks. 
We set aside 30\% of the samples as our test set and used them to evaluate the performance of different codebooks. The final test set accuracies are reported in Table \ref{tab:real_small}. 
Our codebook provides best accuracy on {\em Ecoli} and second-best accuracy on \emph{ Glass} and \emph{Yeast}, thus providing best performance on an average.
\begin{table}[H]
\centering
\caption{Performance of various codebooks on different real-world (small) datasets.}
 \resizebox{\linewidth}{!}{%
\begin{tabular}{|l|c|c|c|c|c|}
\hline
        &  $\mathcal{IP}3$   &   Dense    &  Sparse     &   1-vs-all   &    1-vs-1  \\ \hline
 Glass   &     67.69\%     &  \textbf{75.38\%}   &   67.69\%   &     59.99\%    &     66.15\%    \\ \hline
Ecoli   &     \textbf{90.09\%}     &  87.12\%   &   83.16\%   &     71.28\%    &     77.22\%    \\ \hline
Yeast   &    \textbf{ 51.79\%}     &  50.67\%   &   43.04\%   &     48.20\%    &     \textbf{52.91\%}    \\ \hline
\end{tabular}
}\label{tab:real_small}
\end{table}

We now evaluate the performance of different codebooks on real-world image datasets: MNIST and CIFAR10.

\textbf{MNIST:} We run two set of experiments: In the first set, we use SVMs (with both Linear and Rbf kernel) on PCA-transformed MNIST dataset (using 25 principal components). In the second set, we use binary Convolutional Neural Networks CNNs to train different hypotheses in our codebooks. 
Tables \ref{tab:svm_mnist} and \ref{tab:cnn_mnist} provide the the test set accuracy of different codebooks from both sets of experiments.
\begin{table}[h]
\centering
\caption{Performance of different codebooks using SVM on PCA transformed MNIST dataset.}
\resizebox{\linewidth}{!}{%
\begin{tabular}{|l|c|c|c|c|c|}
\hline
                                & $\mathcal{IP}3$   & Dense    & Sparse     & 1-vs-all & 1-vs-1 \\ \hline
Linear    &   80.37\%      & 75.74\%  & 68.87\%  &  76.82\%     & 92.01\%    \\ \hline
Rbf       &   97.59\%      & 97.5\%   & 79.18\%    &  96.95\%      & 98.01\%    \\ \hline
\end{tabular}
}\label{tab:svm_mnist}
\end{table}
\begin{table}[h]
\centering
\caption{Performance of Different Codebooks  with binary CNN on MNIST dataset.}
\resizebox{\linewidth}{!}{%
\begin{tabular}{|c|c|c|c|c|c|c|}
\hline
 \multirow{2}{*}{$\mathcal{IP}3$} & \multirow{2}{*}{Dense} & \multicolumn{2}{c|}{Sparse}          & \multirow{2}{*}{1-vs-all} & \multirow{2}{*}{1-vs-1} \\ \cline{3-4}
                               &                        & Normalized & Raw &                             &                             \\ \hline
\textbf{98.84}\%                       & 98.8\%                & 95.05\%             & 84.17\%              & 98.65                       & 94.51\%                     \\ \hline
\end{tabular}
}\label{tab:cnn_mnist} 
\end{table}
We observe that in the case of \emph{Linear kernel} our $\mathcal{IP}3$ codebook outperforms all other codebooks except for 1-vs-1, which achieves relatively higher accuracy of around 92\%. This is due to fact that the individual hypotheses of different codebooks (except 1-vs-1 ) are solving much harder problems with \emph{highly non-linear decision boundaries}. On the contrary, 1-vs-1 solves only natural classification problems, where a linear separator can be expected do well. In using \emph{non-linear Rbf-kernel}, both $\mathcal{IP}3$ codebook and 1-vs-1 codebook achieve similar accuracy. On the other hand, when using CNNs our $\mathcal{IP}3$ codebook provides best performance, indicating the benefit of using powerful binary classifiers in our ECOC approach.

\textbf{CIFAR10:} Since running SVMs on this dataset is expensive computationally, we resort to CNNs here. In particular, we use ResNet18 \cite{resnet} as our binary classifier to train the individual hypotheses in different codebooks.
As shown in Table \ref{tab:cifar10}, $\mathcal{IP}3$ achieves the best performance. Note that our experiments on CIFAR10 should be viewed only in terms of evaluating the \emph{relative} performance of different codebooks. We are aware that modern multi-output CNNs have achieved an accuracy of around 95\% (or higher) on CIFAR10 dataset. However, recall that in this work our goal is to highlight the benefit of using ECOCs when working with binary classifiers. 

\begin{table}[h]
\centering
\caption{Performance of Different Codebooks  with binary CNN (ResNet18)  on CIFAR10 dataset.}
\resizebox{\linewidth}{!}{%
\begin{tabular}{|c|c|c|c|c|c|}
\hline
 \multirow{2}{*}{$\mathcal{IP}3$} & \multirow{2}{*}{Dense} & \multicolumn{2}{c|}{Sparse}          & \multirow{2}{*}{1-vs-all} & \multirow{2}{*}{1-vs-1} \\ \cline{3-4}
                               &                        &       Normalized &     Raw   &                             &                             \\ \hline
  \textbf{76.25}\%                    & 75.47\%                & 68.15\%             & 61.53\%              & 71.25\%                       & 68.76\%                     \\ \hline
\end{tabular}
}\label{tab:cifar10}
\end{table}

\subsection{Adversarial Robustness}
We now evaluate the robustness of different codebooks against white-box attacks.\footnote{Different attacks including white-box attacks are defined in supplemental section.}  For further comparison, we also evaluate the robustness of a naturally trained multiclass CNN with our IP-generated codebook in the final layer -- this is somewhat similar to the recent approach in
\cite{Verma_ecoc}.\footnote{Please refer to Supplementary section for more details.} However, note that all our binary hypotheses are  naturally trained, i.e. {\em without any adversarial training}.  We first discuss how to obtain the class probability estimates that are necessary to evaluate the adversarial robustness.

Recall from Sec.~\ref{sec:ecc} the procedure of assigning a class to an input $x$ %
using Hamming decoding. However, this decoding scheme in itself does not provide us with class probability estimates, which are essential for evaluating the robustness of an ECOC-based classifier with respect to white-box attacks~\cite{MadryNet,fgsm}. Particularly, we need probability estimates to compute the adversarial loss function. Furthermore, we need to be able to compute the gradients of the loss-function with respect to input $x$. %

We adopt the procedure of calculating the class probability estimates for general codebooks, as proposed in~\cite{prob_estimates,pairwise_coupling}. After evaluating an input $x$ on each binary classifier, we obtain a probability estimate (or score\footnote{Class scores can be easily converted into probabilities using sigmoid non-linearity.}), denoted $r_l(x)$, for each column $l$ (i.e., binary classifier) in $\mathcal{M}$. Let $I$ denote the set of classes for which $\mathcal{M}(\cdot,l) =1$  and $J$ denote the set of classes for which $\mathcal{M}(\cdot,l)=-1$. Then the class probability estimate for $i \in \{1,\dots k\}$ on an input $x$ is given as follows:
\begin{adjustwidth}{-0.1in}{0.0in}
\begin{equation}
\hat{p}_i(x) =   \sum_{l: \; \mathcal{M}(i,l)=1}r_l(x) + \sum_{l:  \;\mathcal{M}(i,l)=-1}\left(1-r_l(x)\right), \label{eq:non-iter} 
\end{equation}
\end{adjustwidth}
where differentiability with respect to $x$ is maintained.

Using these estimates, we can compute a loss function (e.g, cross-entropy Loss) and then generate white-box PGD-attacks \cite{MadryNet} to evaluate the robustness of the overall classifier. Note that we use the same \textit{differentiable} class scores (or decoding scheme) for both prediction and to generate a white-box attack in order to prevent gradient-obfuscation \cite{obfuscated-gradients, tramer}.
For all our experiments, we work with perturbations based on $l_{\infty}$-norm. In particular, for a given input $x'$, the allowed set of perturbations are given by set:
\begin{center}
$\mathcal{Q}(x')= \{x \in R^d \;\big| \;\;||x-x'||_{\infty} \le \epsilon\;\; ; l \le x \le u \} $.\\ 
\end{center}

\textbf{MNIST:} We run an $l_{\infty}$-norm based 100-step PGD attack with multiple values of $\epsilon$ on different codebooks; Table~\ref{tab:mnist_nominal} summarizes these results. 
In terms of the overall performance, our $\mathcal{IP}3$-generated codebook significantly outperforms all other codebooks except the {\em Dense} codebook. In this codebook, different pairs of codewords have different hamming distances, ranging from 8-14. On the other hand, in $\mathcal{IP}3$, all codeword pairs have identical hamming distance of 10 as result of the max-min objective function \eqref{eq:ip3_obj}. This disparity in performance can therefore be mitigated by incorporating the underlying data distribution (via class pair similarity measures) using the objective function \eqref{ip3:new_obj}. However, note that efficiently computing similarity measures for large image datasets is in itself a research problem. Finally, as we discuss in the next set of experiments, the performance of Dense codebook deteriorates as the data-distribution changes. 

\begin{table}[h]
\centering
\caption{Adversarial Accuracy of Nominally Trained Codebooks on MNIST.}
 \resizebox{\linewidth}{!}{%
\begin{tabular}{|c|c|c|c|c|c|c|}
\hline
                       & $\epsilon = 0.05$ & $\epsilon = 0.1$ & $\epsilon =0.15$ & $\epsilon =0.2$ & $\epsilon =0.25$ & $\epsilon =0.3$ \\ \hline
$\mathcal{IP}3$        &95.46\%  & 83.6\%  & 57.67\% & 29.96\% & 12.99\% &4.81\%    \\ \hline
1-vs-1              & 84.48\% & 59.17\% & 25.57 \%   & 7.91\% & 2.36 \% & 0.66\% \\ \hline
1-vs-All            & 93.64\% & 70.74\% & 30.89\%   &  6.74\% & 1.87\%  & 0.86\%  \\ \hline
Sparse              & 86.12\% & 58.67\% & 22.65\% & 5.4\% & 0.63\%  & 0.01\%    \\ \hline
Dense               & 95.17\% & 84.08\% & 62.95\% & 43.54\% & 28.6\% & 16.34\%  \\ \hline
Multiclass             & 94.35\% & 70.29\% & 21.72\%  & 2.19 \%  & 0.04 \% & 0.0\%   \\ \hline
\end{tabular}
}\label{tab:mnist_nominal}
\end{table}

\textbf{CIFAR10:}  Finally, we evaluate the robustness of different codebooks on CIFAR10 by running  30-step PGD attack; see table \ref{tab:cifar_nominal}. In this case, our $\mathcal{IP}3$ codebook outperforms all other codebooks including Dense codebook. Note that since the data-distribution changed from MNIST to CIFAR10, Dense codebook now shows lower performance than $\mathcal{IP}3$, particularly for larger perturbations of $\epsilon = 4/255$ and $\epsilon=  8/255$. 

\begin{table}[h]
\centering
\caption{Adversarial Accuracy of Nominally Trained Codebooks on CIFAR10.}
\resizebox{0.7\linewidth}{!}{%
\begin{tabular}{|c|c|c|c|}
\hline
                       & $\epsilon = 2/255$ & $\epsilon = 4/255$ & $\epsilon =8/255$  \\ \hline
$\mathcal{IP}3$        & 24.04\%            & 19.24\%            & 16.48\%    \\ \hline
1-vs-1                 & 4.65\%             & 0.11\%             & 0.0 \%   \\ \hline
1-vs-All               & 2.83\%             & 0.14\%             & 0.0\%     \\ \hline
Sparse                 & 5.05\%             & 0.08\%             & 0.0\%     \\ \hline
Dense                  & 24.2\%             & 12.79\%            & 11.63\%   \\ \hline
Multiclass             & 15.46\%            & 2.55\%             & 0.27\%     \\ \hline
\end{tabular}
}\label{tab:cifar_nominal}
\end{table}

Importantly, the adversarial accuracy achieved by our $\mathcal{IP}3$ is by no means trivial as under the exactly same setting other codebooks like 1-vs-1, 1-vs-All, Sparse do not show any robustness. In similar setting, a multi-class CNN of similar network capacity also does not provide any robustness to adversarial perturbations. This highlights the impressive capability of ECOCs to handle adversarial perturbations even though the individual binary hypotheses are all nominally trained.
Our approach provides \emph{robustness-by-design}, and does not make any specific assumptions about the adversary model in the design of codebook. 
\section{Conclusion and Future Work}\label{sec:conclude}
Our computational results validate the merit of our optimal codebook design approach. Importantly, our IP-based formulation achieves small (or zero) optimality gaps while maintaining tractability for reasonable problem sizes. This is possible mainly due the graph-theoretic viewpoint we adopted in applying the edge-clique-cover, which substantially reduced the constraint set of original IP formulation. 
In the nominal setting, our \emph{compact} IP generated codebooks outperform commonly used standard codebooks on most datasets. 

In the adversarial setting, our IP-generated codebooks achieve non-trivial robustness. This is surprising due to three main reasons: (1) We do not employ \emph{any} {adversarial training}; (2) Most other codebooks (except Dense) do not exhibit any robustness even when they use more than twice the number of columns; (3) The robustness that we obtain is not simply because of the large network capacity. To the best of our knowledge, we are the first ones to report that adversarial robustness can be achieved by a careful codebook design approach, while only using nominally trained binary classifiers.

Our results provide guidance for further research in the use of ECOCs for robust classification. We plan to study the effect of robustifying the individual hypotheses.
Another variant would be to use a combination of nominally and adversarially trained hypotheses. We plan to pursue these aspects in our future work.

\newpage
\bibliography{paper}

\end{document}

\onecolumn
\aistatstitle{Supplementary Material}%
\section{Proofs}
\begin{theorem}\label{th:clique}
The feasible space enclosed by the constraints constituting the edges of any clique $\mathcal{C} $ in $\mathcal{G}_p^{inf}$ is same as that enclosed by the single constraint:
\begin{equation}\tag{16}
  \centering
 \displaystyle \sum_{i \in \mathcal{C}} x_i \le 1. \label{eq:single_const}
\end{equation}
\end{theorem}
\setcounter{equation}{20}
\emph{Proof:} We use mathematical induction to show that the result holds for any clique $\mathcal{C}_n$, of size $n\ge 3$. 
Assume that the theorem holds for a clique of size $n-1$.
We know that a clique $\mathcal{C}_n$ (size $n$) contains $n$ distinct cliques $\mathcal{C}_{n-1}^1 , \dots , \mathcal{C}_{n-1}^n$ of size $n-1$ such that $\mathcal{C}_{n-1}^1 \cup \mathcal{C}_{n-1}^2  \dots \mathcal{C}_{n-2}^{n-1} \cup \mathcal{C}_{n-1}^n = \mathcal{C}_{n}$. 
Under induction hypothesis, we can write the following set of $n$ equations:
$$
\begin{array}{cccccccccc}
    x_1     & + & x_2  & + & \cdots   & + & x_{n-1} &    &       &  \le 1  \\
    x_1     & + & x_2  & + & \cdots   & + &         &  + &  x_n  &  \le 1 \\
    \vdots  &   &      &   &  \vdots  &   &         &    & \vdots&  \vdots       \\
    x_1     & + &      &   & \cdots   & + & x_{n-1} &  + &  x_n  &  \le 1  \\
        &   & x_2  & + & \cdots   & + & x_{n-1} &  + &  x_n  &  \le 1
    
\end{array}
$$

Adding these equations, we obtain:
\begin{equation} \label{eq:sum_n}
 \displaystyle x_1 + x_2 +  \cdots  + x_{n-1} + x_n \leq  \frac{n}{n-1}
\end{equation}
For $n \ge 3$, we know the following trivial bound: 
\begin{equation}\label{eq:less_2}
 \frac{n}{n-1} < 2  
\end{equation}

Using \eqref{eq:sum_n} and \eqref{eq:less_2}:
\begin{equation}\label{eq:sum_2}
 \displaystyle x_1 + x_2 +  \cdots  + x_{n-1} + x_n < 2
\end{equation}
Since $x_i \; \in \; \{0 ,1\} \; \forall i \in \{1, \dots, n\}$:\\
\begin{equation}\label{eq:sum_int}
 x_1 + x_2 +  \cdots  + x_{n-1} + x_n \in \mathbb{Z}^{+}\cup\{0\}
\end{equation}

Using \eqref{eq:sum_2} and \eqref{eq:sum_int}: \\
\begin{equation}
 x_1 + x_2 +  \cdots  + x_{n-1} + x_n \leq 1
\end{equation}

Thus, \eqref{eq:single_const} holds for a clique of size $n$. To complete the induction argument, we need to show that the result holds for $n=3$. For $n=3$, we have:
\begin{alignat*}{10}
 x_1 + x_2  & \le 1 \\
 x_1 + x_3  & \le 1 \\
 x_2 + x_3  & \le 1 
\end{alignat*}

Summing the above equations, we get:
\begin{equation*}
 x_1 + x_2+ x_3 \leq 1.5
\end{equation*}
\newpage
Again, since $x_i \; \in \; \{0 ,1\} \; \forall i \in \{1, 2,3\}$:\\
\begin{equation*}
 x_1 + x_2 + x_3 \in {\mathbb{Z}}^{+}\cup\{0\}, 
\end{equation*}
and we conclude that: 
\begin{equation*}
 x_1 + x_2 +  x_3 \leq 1.
\end{equation*}
Therefore, the result also holds for $n=3$. \qed

\begin{corollary} \label{th:edge_cover}
The feasible space enclosed by the constraint set $\mathcal{S}_p^{inf}$ (or its graphical equivalent $\mathcal{G}_p^{inf}$)  in $\mathcal{IP}2$ is same as that enclosed by a much smaller constraint set formed by $\mathcal{ECC}(\mathcal{G}_p^{inf})$.
\end{corollary}
\emph{Proof:} Since the graph $\mathcal{G}_p^{inf}$ does not contain any isolated nodes and loops, and every edge in $\mathcal{G}_p^{inf}$ is covered in atleast one clique, therefore we can write:
 $$\bigcup_{i=1}^k \mathcal{C}_i = \mathcal{G}_p^{inf} .$$
 Also, $\mathcal{G}_p^{inf} \equiv   \mathcal{S}_p^{inf}$, therefore $\{ \mathcal{C}_1, \dots ,\mathcal{C}_k\} \equiv \mathcal{S}_p^{inf}$. \qed

\begin{lemma} \label{lemma:1}
Suppose $ \mathcal{G}_1, \dots \mathcal{G}_m$ are edge-disjoint subgraphs of  $\mathcal{G}_p^{inf}$, such that:
\begin{enumerate}
 \item $ \mathcal{G}_i \cap  \mathcal{G}_j = \phi \;\; \forall \;  i,j \in \{1, \dots, m\}^2| i< j $
 \item $ \bigcup_{i=1}^m \mathcal{G}_i = \mathcal{G}_p^{inf}$
\end{enumerate}
The union of the edge clique covers of individual subgraphs $ \mathcal{G}_1, \dots \mathcal{G}_m$ is a valid edge clique cover of $\mathcal{G}_p^{inf}:$ 
$$\;\; \bigcup_{i=1}^m \mathcal{ECC}(\mathcal{G}_i) = \mathcal{ECC}(\mathcal{G}_p^{inf}).$$ 
\end{lemma}

\emph{Proof:} Recall form the definition of Edge Clique Cover, a set of cliques is a valid edge-clique-cover of a given graph, if the following two requirements are satisfied by the clique set:
\begin{enumerate}[I]
 \item Every edge of the graph is covered in atleast one clique.
 \item No clique is completely contained in another clique.
\end{enumerate}

Consider the following arguments:
\begin{enumerate}
 \item For any $i \in \{1,\dots m \}$, $\mathcal{ECC}(\mathcal{G}_i)$ is a valid edge-clique-cover for subgraph $\mathcal{G}_i$. 
 \item Every edge in $\mathcal{G}_p^{inf}$ is covered in atleast one subgraph as $ \bigcup_{i=1}^m \mathcal{G}_i = \mathcal{G}_p^{inf}$, therefore every edge in $\mathcal{G}_p^{inf}$ is contained in atleast one of the cliques in the set: $\bigcup_{i=1}^m \mathcal{ECC}(\mathcal{G}_i)$. Thus requirement I is satisfied.
 \item For a clique to be completely contained in another clique, there should be atleast one common edge between any two distinct subgraphs $\mathcal{G}_i$ and $\mathcal{G}_j$. Since, the subgraphs are edge-disjoint, i.e.  $\mathcal{G}_i \cap  \mathcal{G}_j = \phi$, therefore no clique can be completely contained in another clique. Thus requirement II is also satisfied.  \qed 
\end{enumerate}

\section{A sample exhaustive code}
\begin{table}[H]
\centering
\caption{Exhaustive code (all possible valid columns) for $k =5$}
\label{tab:exhaustive}
\begin{tabular}{|c|c|c|c|c|c|c|c|c|c|c|c|c|c|c|c|}
\hline
\multirow{2}{*}{Classes} & \multicolumn{15}{c|}{Codewords}                                 \\ \cline{2-16} 
                         & f1 & f2 & f3 & f4 & f5 & f6 & f7 & f8 & f9 & f10 & f11 & f12 & f13 & f14 & f15 \\ \hline
1  & 1   &  1   &  1   &  1   &  1   &  1   &  1   &  1   &  1   &  1   &  1   &  1   &  1   &  1   &  1    \\  \hline   
2  & -1   &  -1   &  -1   &  -1   &  -1   &  -1   &  -1   &  -1   &  1   &  1   &  1   &  1   &  1   &  1   &  1   \\ \hline  
3  & -1   &  -1   &  -1   &  -1   &  1   &  1   &  1   &  1   &  -1   &  -1   &  -1   &  -1   &  1   &  1   &  1   \\ \hline 
4  & -1   &  -1   &  1   &  1   &  -1   &  -1   &  1   &  1   &  -1   &  -1   &  1   &  1   &  -1   &  -1   &  1   \\ \hline 
5  & -1   &  1   &  -1   &  1   &  -1   &  1   &  -1   &  1   &  -1   &  1   &  -1   &  1   &  -1   &  1   &  -1   \\ \hline 
\end{tabular}
\end{table}

\section{Dense/Sparse Codes}\label{sec:rand_codes}
\textit{Random codes} is another way of generating codebooks as proposed in \cite{allwein}. Here, authors propose generating $10000$ matrices, whose entries are randomly selected. If the elements are chosen uniformly at random from $\{+1,-1\}$, then the resulting codebooks are called \textbf{dense codes} and if the elements are taken from $\{-1,0,+1\}$ then the resulting codebooks are called \textbf{ sparse codes}. In sparse codes, $0$ is chosen with probability $1/2$ and $\pm1$ are each chosen with probability $1/4$. Out of the $10,000$ random matrices generated, after discarding matrices which do-not constitute a valid codebook, \emph{the one with the largest minimum Hamming distance among rows is selected.}
Note that since out of the $10,000$ matrices the one with the \emph{largest minimum hamming distance} is selected, therefore despite the matrices being generated randomly, the final codebook \emph{can have high row-separation.} 

\section{Details about Real-world datasets(Small/Medium)}
In table \ref{tab:data_char} we provide details about {\em Glass}, {\em Ecoli} and {\em Yeast} datasets taken from the UCI repository \cite{uci}.
\begin{table}[h!]
\centering
\caption{Real-world Dataset Characteristics}
\begin{tabular}{|l|c|c|c|} \hline
        & \# of samples & \# of features & \# of classes (k)\\ \hline
 Glass   &     214       &      9         &       6        \\ \hline
Ecoli   &     336       &      7         &       8        \\ \hline
Yeast   &     1484      &      8         &       10        \\ \hline
\end{tabular}\label{tab:data_char}
\end{table}

\section{Class Probability Estimates }

In section 5.2 under Adversarial Robustness, we  discussed how class probability estimates enable us to estimate the adversarial robustness of ECOC based classifiers using white-box attacks. For binary codebooks we obtain class probability estimates using the procedure from  ~\cite{prob_estimates,pairwise_coupling}.
After evaluating an input $x$ on each binary classifier, we obtain a probability estimate, denoted $r_l(x)$, for each column $l$ (i.e., binary classifier) in $\mathcal{M}$. Let $I$ denote the set of classes for which $\mathcal{M}(\cdot,l) =1$  and $J$ denote the set of classes for which $\mathcal{M}(\cdot,l)=-1$. Then the class probability estimate for $i \in \{1,\dots k\}$ on an input $x$ is given as follows:

\begin{equation}\tag{20}
\hat{p}_i(x) =   \sum_{l: \; \mathcal{M}(i,l)=1}r_l(x) + \sum_{l:  \;\mathcal{M}(i,l)=-1}\left(1-r_l(x)\right), \label{eq:non-iter} 
\end{equation}
where differentiability with respect to $x$ is maintained.

The above estimates work well for binary codes, however we need to be careful for ternary (or sparse) codes. For ternary codes,  hypotheses which have zero (for a particular class) do-not contribute to the above sum in \eqref{eq:non-iter}. Therefore due to zero entries, estimates for different classes  can significantly vary in relative magnitude. This can be easily fixed by simple normalization. Raw estimates in \eqref{eq:non-iter} can be normalized as follows:
\begin{equation}
\hat{p}^{*}_i =     \frac{1}{ \sum_{l=1}^{L} \mathbbm{1}_{\big\{ \;\mathcal{M}(i,l) = 1 \vee \mathcal{M}(i,l) = -1 \; \big\} } }    \Big( \;\;  \sum_{l: \; \mathcal{M}(i,l)=1}r_l(x) + \sum_{l:  \;\mathcal{M}(i,l)=-1}\left(1-r_l(x)\right) \;\; \Big) ,
\end{equation}

where $\mathbbm{1}_{ \{\pi\} }$ is the indicator function which evaluates to $1$ when the predicate $\pi$ is true and $0$ otherwise.

\begin{figure}[H]
    \centering
    \includegraphics[scale = 0.4]{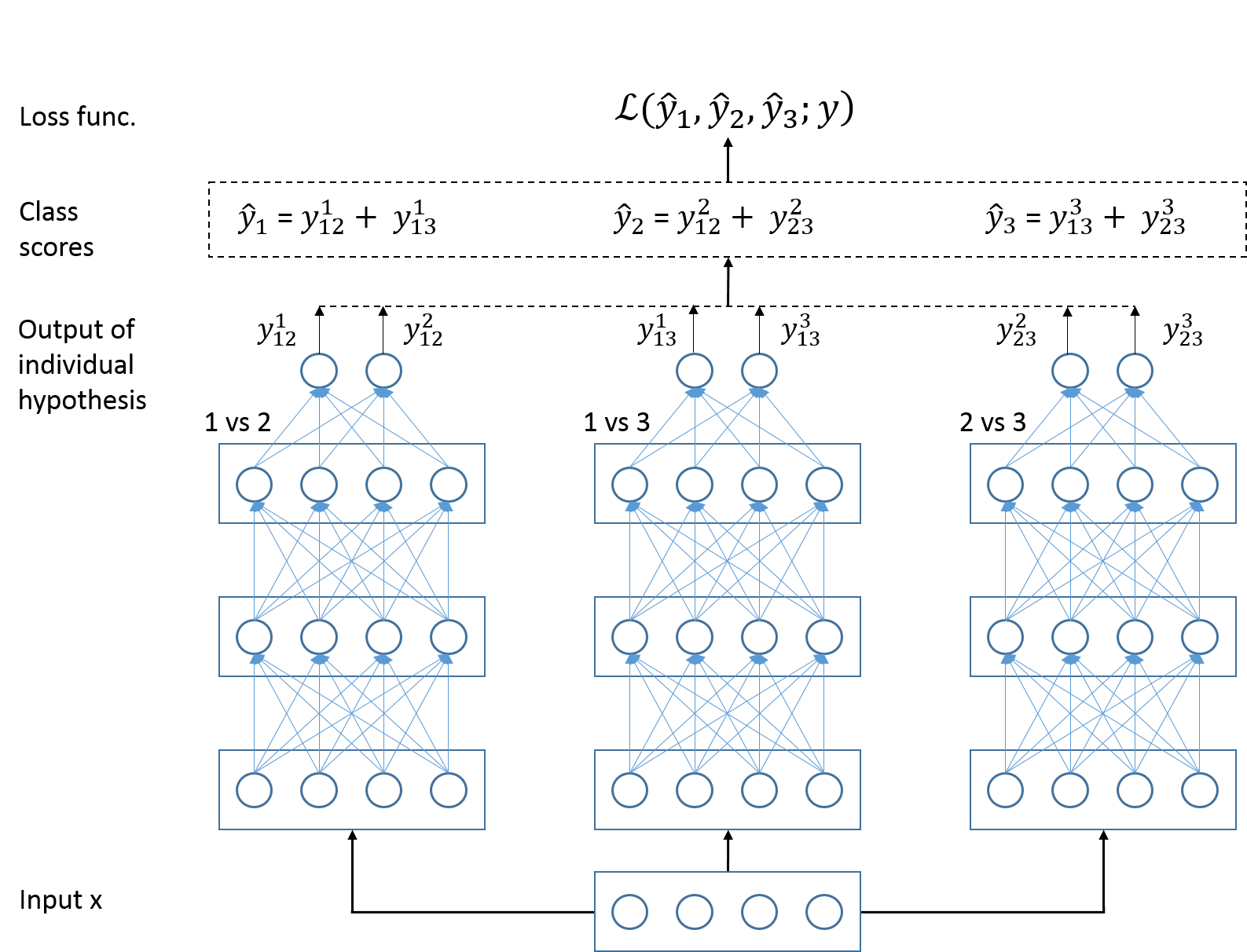} 
    \caption{Combining output of individual hypotheses of 1-vs-1 to generate class scores while maintaining differentiability.}
    \label{fig:all-pairs-1}
\end{figure}

In figure \ref{fig:all-pairs-1}, we show how these estimates can be computed for 1-vs-1 codebook, when working with binary deep neural networks. Since, each row in 1-vs-1 has the same number of zeros therefore normalization is not necessary.

\section{ Adversarial Accuracy and Different types of Attacks}
For ECOC based classifiers, evaluation of  natural or clean accuracy over an example (generally from test-set) is straightforward, and can be easily done either by using a decoding scheme such as Hamming decoding or by calculating class probability estimates and choosing the class with the highest probability.

\iffalse
 However, calculation of adversarial accuracy is not straightforward as it involves solving a \textit{non-concave} constrained maximization problem. Recall that since we are in the domain of \textit{worst-case} analysis, if the model correctly predicts over \textit{all} allowed perturbations in $\mathcal{S}$ or equivalently if the model correctly predicts for the \textit{worst-case} perturbation, \textit{only then} we can correctly 
conclude that the model is robust for that particular example. Due to highly non-convex nature of neural networks, solving the non-concave constrained maximization problem remains the main challenge in correct evaluation of the adversarial accuracy. 
\fi
We now mathematically define the problem of evaluating the adversarial accuracy using class probability estimates.
Suppose $c$ be the true class associated with a given input $x'$ and let $i \in \{1,\dots,k \}/\{c\}$  be the target class for which the attacker is trying to generate an adversarial perturbation. Attacker aims to  solve the following non-convex problem: 

\begin{equation}
f^{*}(x') = \displaystyle \max_{\delta: \; x'+\delta \;\in\; \mathcal{Q}(x') } \;\; \hat{p}_i(x'+\delta)- \hat{p}_c(x'+\delta) 
\label{eq:attack}
\end{equation}

In \eqref{eq:attack}, set $\mathcal{Q}(x')$ for $l_\infty$-norm based perturbations is given as follows:
\begin{center}
$\mathcal{Q}(x')= \{x \in R^d \;\big| \;\;||x-x'||_{\infty} \le \epsilon\;\; ; l \le x \le u \} $.\\ 
\end{center}

For a valid\footnote{An adversarial perturbation $\delta$ does not necessarily need to be the $\arg \max$ of \eqref{eq:attack}} adversarial perturbation $\delta$, the objective function value of \eqref{eq:attack} would be strictly positive for some target class $i$. 
\iffalse
Note that for the purposes of calculating adversarial accuracy, we do not really care about the \textit{exact} maximizer of \eqref{eq:attack}. Instead we are mainly interested in determining one of the following two things:
\begin{enumerate}\vspace{-0.3cm}
 \item Can we find a feasible solution of \eqref{eq:attack} for which the objective function value is positive?\vspace{-0.4cm} (Category \rom{1})
 \item Whether the optimal value of the maximizer is strictly positive or not? (Category \rom{2})
\end{enumerate}
\fi
Different attacks such as black-box and white-box attacks attempt to solve the above outlined problem \eqref{eq:attack} under different settings (or threat model). 

In \emph{black-box} setting, only the output of the classifier i.e. the class probabilities or score of each class is known to the attacker. No model information is available to the attacker, i.e. the network architecture and the weights of the network. In this setting, since only class probability estimates are available, therefore analytical computation of gradients is not possible. The problem is generally solved using off-the-shelf black-box optimizers comprising of  heuristics based algorithms such as Particle Swarm Optimization (PSO), Genetic Algorithms (GAs) etc. However, given the efficacy of gradient based attacks, one can also try to compute an estimate of the gradient and then use this estimate to run gradient-based attacks, for details see \cite{black-box}. SPSA proposed in \cite{Spall92} is another black-box optimization method which is based on gradient estimation.

In \emph{white-box} setting, the class probability estimates along with the model architecture and weights are known to the attacker. White-box setting can also be referred to as complete information setting. In white-box setting, the projected gradient descent or the PGD-attack proposed in \cite{MadryNet} has emerged as one of the strongest known attack.
Another popular gradient based attack known as Fast-Gradient-Sign method (FGSM) was proposed in \cite{fgsm}. FGSM can be viewed as simply a single step PGD attack and therefore is a much weaker attack in comparison to PGD-attack.

Given the non-concave nature of the problem \eqref{eq:attack}, the above attacks do not provide any guarantee in terms of finding the optimal solution, and mainly aims at finding a feasible solution to \eqref{eq:attack} with positive objective function value. 
If these attacks fail in generating an adversarial perturbation (especially if the attack is weak), we conclude that the model is robust against that particular attack. Therefore, to estimate the adversarial robustness accurately it is important to evaluate against strongest possible attack.

\section{Comparison with Multiclass CNN}%
In section 5.2 we compared the adversarial robustness of our IP generated codebook $\mathcal{IP}3$ with other standard codebooks such as 1-vs-1,1-vs-All, Sparse and Dense codes. We reported our results on MNIST and CIFAR10 datasets in table 8 and 9 respectively. We note that our IP generated codebook achieves non-trivial robustness \emph{without any adversarial training.}
On CIFAR10, our codebook outperforms all other standard codebooks, achieving an adversarial accuracy of $\sim 16\%$ with $\epsilon =8/255$. However, given that we are combining the output of $20$ binary classifiers, each of which is a ResNet-18, a natural question arises:
\begin{center}
\textit{Is network capacity (of the overall classifier) the main reason for this robustness?} 
\end{center}

Recall that to evaluate the robustness we combine the outputs of each of the hypotheses (individually trained before) using our IP generated codebook and form a multi-class classifier. Figure \ref{fig:all-pairs-1} shows this for 1-vs-1 codebook for 3 classes.  We then do a PGD based evaluation of the resulting multiclass classifier.  To investigate the role of network capacity, we now in the same manner, combine 20 \emph{untrained} hypotheses (ResNet-18) to form a multi-class classifier (say $\mathcal{F}(x)$). We now nominally train this 10-class classifier $\mathcal{F}(x)$ \emph{end-to-end} using the entire training set. $\mathcal{F}(x)$ has exactly the same network architecture and capacity as our multiclass ECOC based classifier resulting from our IP generated codebook.

We now evaluate the adversarial accuracy of $\mathcal{F}(x)$ using the same PGD attack which we used for different codebooks including our IP generated codebook. We report our results in the last row of table 8 and 9 with type as \emph{Multiclass}. The lack of robustness of $\mathcal{F}(x)$ or \emph{Multiclass} shows that network capacity alone in itself is not the reason for robustness of IP generated ECOC based classifier.

Finally, we note that since the individual untrained hypotheses are combined using a codebook in the final layer, therefore $\mathcal{F}(x)$ is similar to the approach taken in \cite{Verma_ecoc}.

\section{Estimating Error-Correlation between individual hypotheses of a codebook}
In our discussion in section 3, we highlighted that in communicating over a noisy channel, Error-Correcting Codes are powerful only when the errors made due to noise are random. For classification setup like ours, this implies that any two hypotheses (or classifiers) should not make errors on the same inputs. To avoid this, we ensured large column separation in our IP formulation. 
However,  we may still end up with hypotheses whose final predictions (or errors) are correlated. Therefore, measurement of such pairwise correlations between hypotheses can provide us with insights to better understand the final performance of a particular codebook. Moreover, it also will provide us with corroborative evidence to the fact that correlation between hypotheses should be avoided.

Assuming that we have already trained each of our individual hypotheses for a given codebook. Also, let $N_{test}$ denote the number of images in our test-set. For every binary classifier (corresponding to a column) in the codebook, we can compute the $0$-$1$ loss  for all images in the test set so that we have a vector $h_l \in \{0 ,1\}^{N_{test}} \; \forall \; l \in \{1, \dots, L\}$.  We  can now compute the error-correlations between these binary vectors $h_i \; \& \;h_j \;\forall \; (i,j)\; \in \; \{1, \dots, L\}^2$. This can be represented in a $L \times L$ matrix, which we will refer to as the correlation matrix (denoted as $\mathcal{P}$) in our subsequent discussion. We propose the following measure:

\begin{equation}
 \displaystyle \mathcal{P}_{i,j} = \frac{ \sum_{n=1}^{N_{test}}  \mathds{1}\{ h_i[n] = 1 \land h_j[n] = 1\} }{\sum_{n=1}^{N_{test}}  \mathds{1}\{ h_i[n] = 1 \land h_j[n] = 1\} + \sum_{n=1}^{N_{test}} \mathds{1}\{ h_i[n] = 0 \land h_j[n] = 0\} }, \label{error-and}
\end{equation}

where $\mathds{1}\{ \pi \}$ is the indicator function which evaluates to $1$ when the predicate $\pi$ is true and $0$ otherwise.

The above measure \eqref{error-and} accounts for both the correct and incorrect predictions made by individual hypotheses. The magnitude of this error-correlation measure (or the values in the error-correlation matrix $\mathcal{P}$)  will help us in understanding the accuracy of the overall classifier or codebook.

\begin{figure}[h]
 \centering
    \begin{subfigure}{0.4\linewidth}
     \includegraphics[scale = 0.4]{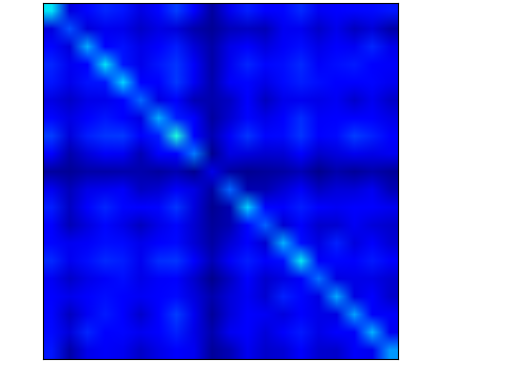}
     \caption{Natural (Accuracy: 76.25 \%) \hspace{2.2cm}  }
    \end{subfigure}
   \begin{subfigure}{0.4\linewidth}
     \includegraphics[scale = 0.4]{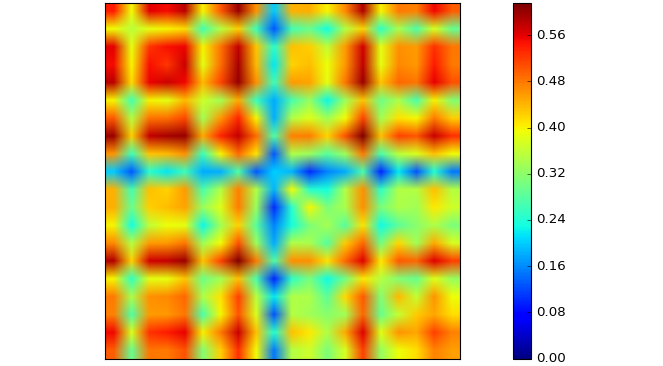}
     \caption{ Adversarial (Accuracy:16.48\%) }
    \end{subfigure} 
    \caption{Error-Correlation matrices estimated using the hypotheses of the IP-generated codebook  on natural and adversarial images of CIFAR10 dataset.} \label{fig:err_mat}
\end{figure}

We estimate the error-correlation matrix using the natural images from CIFAR10 dataset for the nominally trained hypotheses of our IP generated codebook. For the same hypotheses, we also estimate the error-correlation matrix using the adversarial images obtained from the PGD-attack with $\epsilon =8/255$ on the overall classifier. We plot both the matrices in figure \ref{fig:err_mat}.

From figure \ref{fig:err_mat}, we note that the error-correlation values on the natural and adversarial dataset differ by almost an order of magnitude. On natural images, much higher accuracy is achieved as the error-correlation is low, while on adversarial images higher error-correlation values result in lower accuracy. Therefore for higher accuracy, error-correlation between hypotheses should be avoided.   

\clearpage
\section{Various IP Generated Codebooks}\label{sec:ip_codes}

\begin{figure}[H]
\centering
\includegraphics[scale = 0.6]{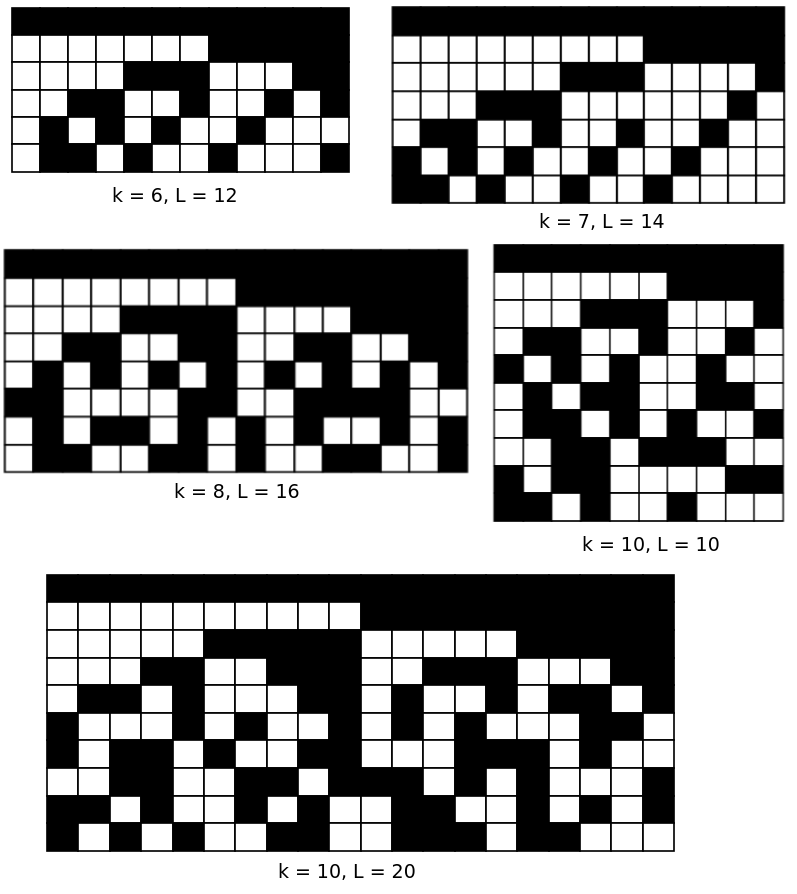} 
\end{figure}

\bibliography{supp}